# Word Spotting in Cursive Handwritten Documents using Modified Character Shape Codes

Sayantan Sarkar

Department of Electrical Engineering, NIT Rourkela

sayantansarkar24@gmail.com

**Abstract.** There is a large collection of Handwritten English paper documents of Historical and Scientific importance. But paper documents are not recognised directly by computer. Hence the closest way of indexing these documents is by storing their document digital image.

Hence a large database of document images can replace the paper documents. But the document and data corresponding to each image cannot be directly recognised by the computer.

This paper applies the technique of word spotting using Modified Character Shape Code to Handwritten English document images for quick and efficient query search of words on a database of document images. It is different from other Word Spotting techniques as it implements two level of selection for word segments to match search query. First based on word size and then based on character shape code of query. It makes the process faster and more efficient and reduces the need of multiple pre-processing.

## 1  Introduction:

Word Spotting is a technique of word recognition in document images without recognizing the characters that constitute the word. It is based on gross classification of characters according to their shape.[2]

Earlier word recognition in printed or handwritten document images was based of OCR (Optical Character Recognition). The accuracy of OCR though was better for printed document images, was sharply reduced for cursive handwritten document images. [5]

Thus, the new technique of word recognition was developed known as word spotting. It has high tolerance for poor image quality, tenability to the lexical content of the documents to which it is applied and high speed of operation.[5]

## 2     Approach:

The proposed method of word spotting can be basically categorised into following stages:

1. Word Segmentation from the document image
2. Indexing individual words segments
3. Efficient Shape Code extraction of the segmented words
4. A user interactive search interface that takes query for Information Retrieval.

The approach is applied to Cursive Handwritten English document images as mentioned.

## 3     Pre Processing- Word Segmentation

In case of handwritten documents, there are no general assumptions to be made. Font size can vary along the document over a considerable range. No assumptions can be made such that two similar characters in the same document can be similar or that the lines are perfectly horizontal in respect to the edges of the document is invalid.[4]

Still for the sake of analysis we take 2 basic assumptions that hold true to a considerable extent.

1. The handwriting though cursive, is legible and has a slant less than 30 degrees to the vertical.
2. The lines in the document are not exactly horizontal, but still two consecutive lines have inter line gap to separate them.

Considering the above mentioned assumptions to be true, a RGB image of the handwritten document is taken as input. It is converted to Binary image and then threshold at a near 0.5 value into two levels 0(black) and 1(white).[3]

### 3.1    Line Segmentation:

As it is already assumed that though the lines of the documents may not be perfectly horizontal, they will not have so much of skew such that there is no inter line gap. Hence the lines are assumed to be near horizontal.

For the document a histogram projection profile of the number of black pixels in a row is created for every row of the image. As both the font size and inter line gap is variable over a large range here, there are less chances of the projection profile to have any kind of pattern.[4]

Thus, irrespective of number, all the consecutive rows that have a high value at the projection profile are grouped as line rows. And all the rows that have a near zero value at the projection profile is considered as inter line gap.[3]

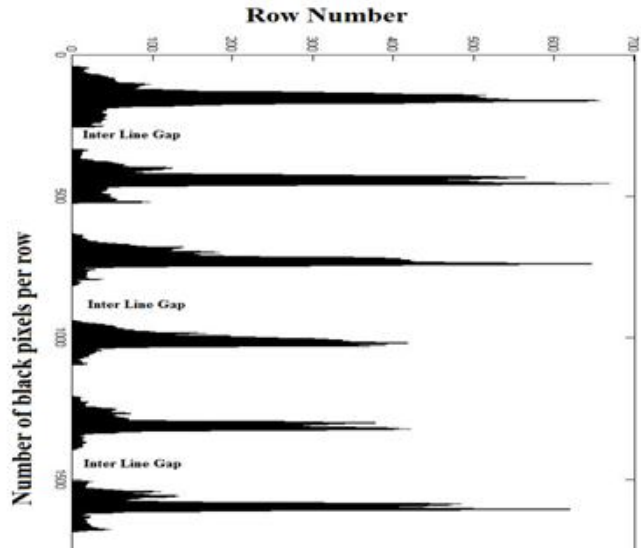

**Fig. 1.** Projection Profile of a handwritten document image highlighting the lines and Inter Line Gap

As its assumed there is inter line gap between all the lines. Thus, all the lines in the document will be grouped individually and extracted

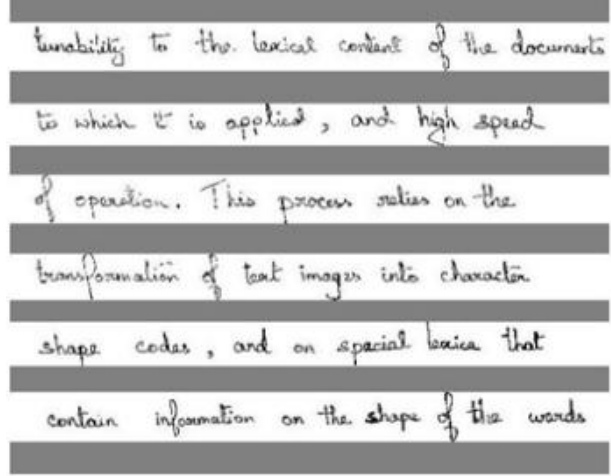

**Fig. 2.** Cursive Handwritten English Document Image with segmented lines

### 3.2 Word Segmentation:

For word segmentation, a segmented line is selected. As the font size of that line is not specified, the height of the line is measured which corresponds to its font size.

For this segmented line a vertical projection profile is generated counting the number of black pixels for each column in the line along its height[3]. As the document is written in cursive handwriting, there should ideally be no Inter Character Gap as all the characters in a given word should be joined.[4]

But in actual test cases though most of the characters are joined in a word sometimes few characters, especially if they are of Upper Case are not joined to its next character[8]. So in the projection profile the number of consecutive columns that have near zero value on the projection profile is calculated. If they are in the order of 0.2*Font Size (experimentally analyzed for variable font sizes), they are considered Inter Character Gap. Else they are classified as Inter Word Gap and neglected.

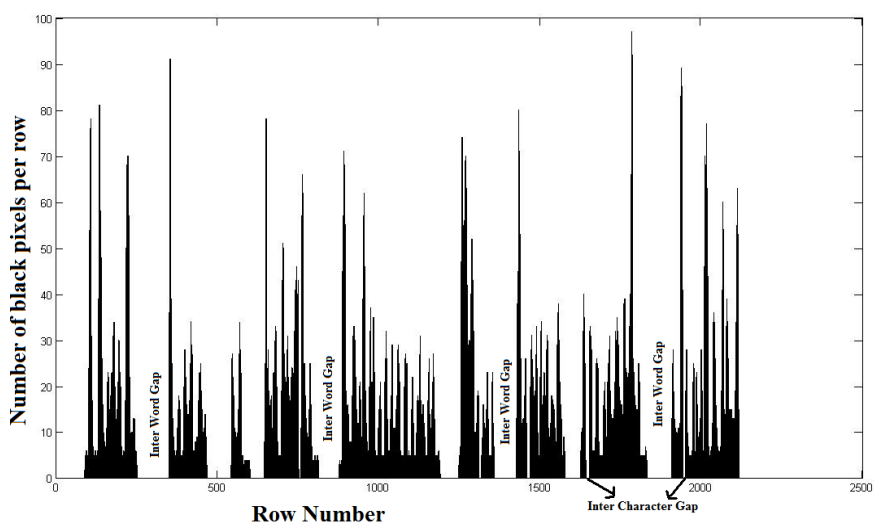

**Fig. 3.** Projection Profile of a segmented line in a Cursive Handwritten Document highlighting Inter Word Gaps and Inter Character Gaps

This helps in proper segmentation of the words in the line. The process is repeated for all the lines to extract the segmented words from the document image.[3]

**Sayantan Sarkar**

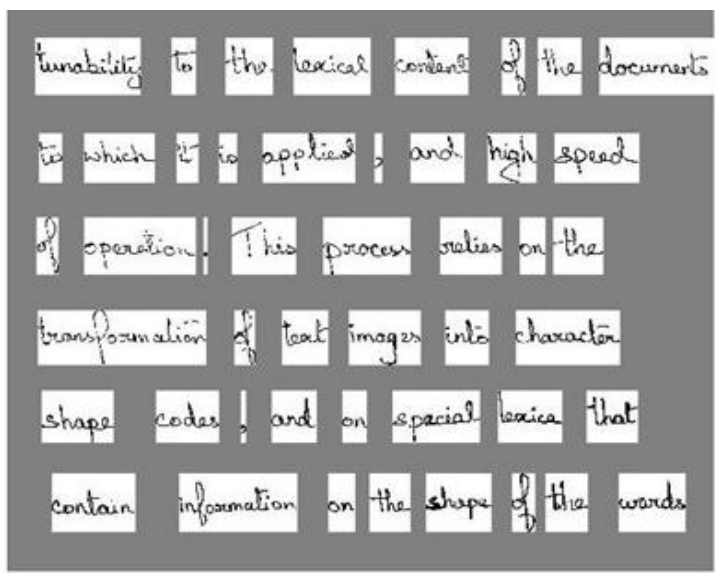

**Fig. 4.** Cursive Handwritten Document Image with segmented words

## 4      Pre Processing: Indexing of Segmented Words

In case of handwritten documents the segmented words are represented by their top left pixel coordinates and bottom right pixel coordinates. They can have variable height and variable word length.

$$Y2 - Y1 = H \ (Word\ Height) \quad (1)$$

$$X2 - X1 = L \ (Word\ Length) \quad (2)$$

Taking word font size K as in that of single font size document as the reference standard, all the word lengths are converted to word lengths for font size K, as the word lengths are proportional to the word height.

L' is the new word length given by:

$$L' = \left(\frac{K}{H}\right) * L \quad (3)$$

L' is standard word length corresponding to each handwritten word independent of the font size of the word.

The words can be generally classified in following basic categories:

| Word Size | L (Pixels) for given test font size |
|---|---|
| Very Small | <80 |
| Small | 80-240 |
| Medium | 240-320 |
| Large | 320-480 |
| Very Large | >480 |

**Table 1.** Classification of Word Size according to their Pixel length (L)

Hence the segmented words are indexed in an array according to their word sizes.

Thus for searching a word in an archive of document image we can just go through the database of segmented words rather than going through the whole image. This helps in quick query search as discussed later.

## 5 Efficient Shape Code Extraction from word segments

As words were successfully segmented earlier for Cursive Handwritten Document Images, instead of a document image we have a collection of indexed segmented words arranged according to their word sizes.[1]

The gross classification of the characters for their Character Shape Code is solely dependent on the top segment and the bottom segment of the characters.[1]

1. *Ascenders:* It is a connected component in the top segment of a character.
2. *Descenders:* It is a connected component in the bottom segment of a character.

For each of the word segment a histogram projection profile is generates which counts the number of black pixel along each column of the word segment. Unlike printed document images, the cursive handwritten words do not have Inter Character Gaps in a word. This is because the characters are connected to each other by a single curved line.[2]

Hence in the histogram there will be no near zero values. So, the earlier discussed methods of character segmentation fails here.

But it is observed that the Inter Character Regions of the word though does not have a zero histogram value, but do have similar histogram values which corresponds to the minimum value for that histogram. Hence the word can be further segmented at the regions which have near minimum value in the histogram.

Such segmentation leads to over-segmentation in most of the words

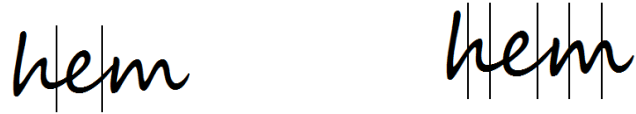

**Fig. 5.** Expected Segmentation and Actual Segmentation of Cursive Words

It is caused due to the fact that few characters in English Alphabet has similar projection profile in their spatial structure to that of the character connectors in cursive words.

This cause multiple segmentation of a single character as the program confuses it with Inter Character Gap. It can be accommodated by modifying the Character Shape Code rules in appropriate manner.

## 6 Modified Character Shape Code

Character Shape Code (CSC) is a process on encoding the gross features of character image ignoring the high spatial detailed features as in case of Optical Character Recognition (OCR)[6]. This is helpful in our analysis because extracting features that are necessary for OCR is highly time consuming.[1]

In case of Cursive Handwritten documents the over segmentation of characters sometimes causes 1 character to be segmented into 2 or 3 regions. Hence the Modified Character Shape Coding takes that into account.[9]

Thus instead of each character, for each of the segmented regions a Shape Code is assigned depending on its Ascenders and Descenders, and then the character is represented by the combination of shape codes, one for each of its segmented regions.[9]

For one character image, the shape code can be 1 or 2 or 3 character long.

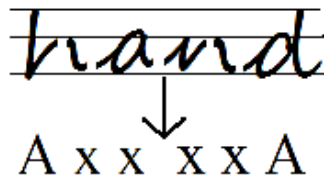

**Fig. 6.** Modified segmentation of word and corresponding Character Shape Code

| Character Shape Code | Character Image | Number of segments |
|---|---|---|
| A | A B C D E F G I J K O P Q R S T X Y Z b d k l t | 1 |
| A A | H M N U V W | 2 |
| A x | L h | 2 |
| X | a c e i o s x z | 1 |
| G | g p q j f | 1 |
| x x | n r u v | 2 |
| x g | Y | 2 |
| x x x | m w | 3 |

**Table 2.** Modified Character Shape Coding for character images

Hence after segmentation of the word segment into regions a corresponding Character Shape Code is generated on the basis of following rules[1]:

1. Regions with ascenders but no Descenders (A)
2. Regions with Descenders but no Ascenders (g)
3. Regions with neither Ascenders nor Descenders(x)

After the Character Shape Code (CSC) is generated for all the character images in the document image, it is necessary to collect them to generate words which will further be used for word spotting. Hence, the words generated by aggregating the Character Shape Codes are known as the Word Shape Tokens (WST) for a particular word image in the document image.[7]

Thus for a given series of word images in a line, it is replaced with a series of Word Shape Tokens representing the words in a gross classified state.[7]

## 7    User Interactive Query Retrieval

Thus after a character string input is taken its size is measured and accordingly from the database of word segments only those words are selected that has a similar size to that of the input query +/- 1 character. For all this word segments, a Word Shape Token is generated but coding the character segments constituting the word with the help of Modified Character Shape Code as discussed earlier. A selected database of Word Shape Tokens is generated for the document image.

**Sayantan Sarkar**

Also the given search query is in form of a character string. Thus, each character in the string is replaced with their corresponding Modified Character Shape Code according to Table 4(Modified Character Shape Coding for character images.) to generate WST.

Most of the time the WST representing the query word does not matches exactly with the WST of the word segments same as the query words. This results in ambiguous output results. But this can be removed to a great extent using the concept of Levenshtein Distance.[5]

The threshold is best decided by trial and test method for a given form of handwriting, but for most cases (including test images) 2.5 is a good estimate of the threshold Levenshtein Distance.

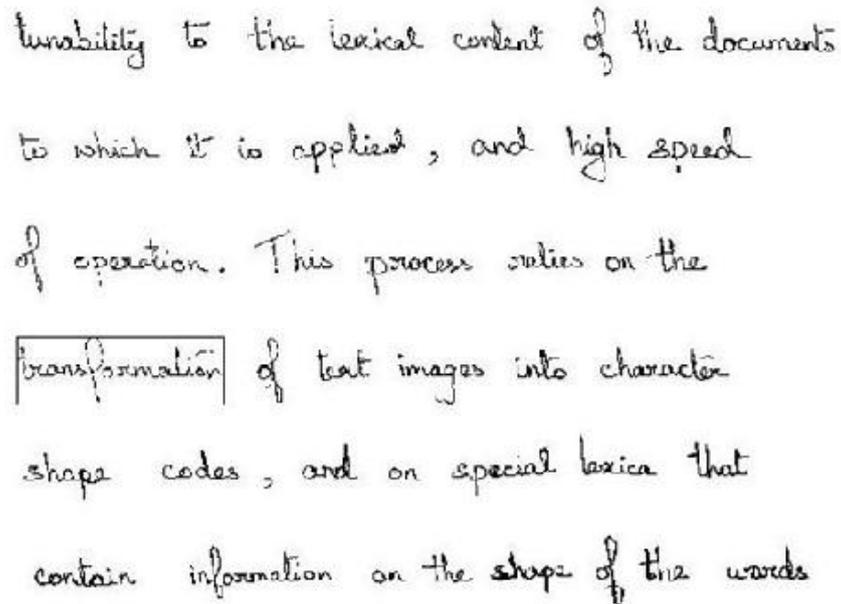

**Fig. 7.** Results for search query "transformation" in a Cursive Handwritten Document Image

## 8   Limitations

Word Spotting method discussed above has less processing time compared to ordinary Optical Character Recognition methods. Also it gives high accuracy results for Cursive Handwritten Documents. Thus, it is highly adaptable to moderate spatial noise and skewing.

But in this method discussed, for search query words that have 4 or less than 4 characters, the corresponding Word Shape Tokens generated are mostly not unique for that word. Hence it identifies few extra words in the document, reducing its accuracy.

For search query words have more than 5 characters, Word Shape Tokens are generally unique for the test set of documents image and the results are highly accurate.

## 9      Conclusion

This paper successfully modifies and implements the Modified Character Shape Code algorithm for cursive handwritten document images for word spotting. This is an up gradation to the well established method of Character Shape Coding developed by A.L.Spitz that usually failed to achieve good accuracy for handwritten documents.

This paper also reduces the operation time of query search of indexed database of word segments implementing size classified query search, which is specifically helpful for query search in large word segment database(like historical books and journals).